\documentclass[10pt,twocolumn,letterpaper]{article}

\usepackage{btas}
\usepackage{times}
\usepackage{epsfig}
\usepackage{graphicx}
\usepackage{amsmath}
\usepackage{amssymb}
\usepackage{subcaption}
\usepackage{array}
\usepackage{url}
\newcolumntype{L}[1]{>{\raggedright\let\newline\\\arraybackslash\hspace{0pt}}m{#1}}
\newcolumntype{C}[1]{>{\centering\let\newline\\\arraybackslash\hspace{0pt}}m{#1}}

\btasfinalcopy 

\ifbtasfinal\fi
\begin{document}

\title{Synthetic Iris Presentation Attack using iDCGAN}

\author{Naman Kohli$^1$, Daksha Yadav$^1$, Mayank Vatsa$^{1,2}$, Richa Singh$^{1,2}$, Afzel Noore$^2$ \\
$^1$West Virginia University, USA, $^2$IIIT Delhi, India \\
\{naman.kohli, daksha.yadav, afzel.noore\}@mail.wvu.edu, \{mayank, rsingh\}@iiitd.ac.in
}

\maketitle
\thispagestyle{empty}

\begin{abstract}

Reliability and accuracy of iris biometric modality has prompted its large-scale deployment for critical applications such as border control and national ID projects. The extensive growth of iris recognition systems has raised apprehensions about susceptibility of these systems to various attacks. In the past, researchers have examined the impact of various iris presentation attacks such as textured contact lenses and print attacks. In this research, we present a novel presentation attack using deep learning based synthetic iris generation. Utilizing the generative capability of deep convolutional generative adversarial networks and iris quality metrics, we propose a new framework, named as iDCGAN (iris deep convolutional generative adversarial network) for generating realistic appearing synthetic iris images. We demonstrate the effect of these synthetically generated iris images as presentation attack on iris recognition by using a commercial system. The state-of-the-art presentation attack detection framework, DESIST is utilized to analyze if it can discriminate these synthetically generated iris images from real images. The experimental results illustrate that mitigating the proposed synthetic presentation attack is of paramount importance.

\end{abstract}

\section{Introduction}

\begin{figure}[t]
 \centering
 \includegraphics[width=1.0\linewidth]{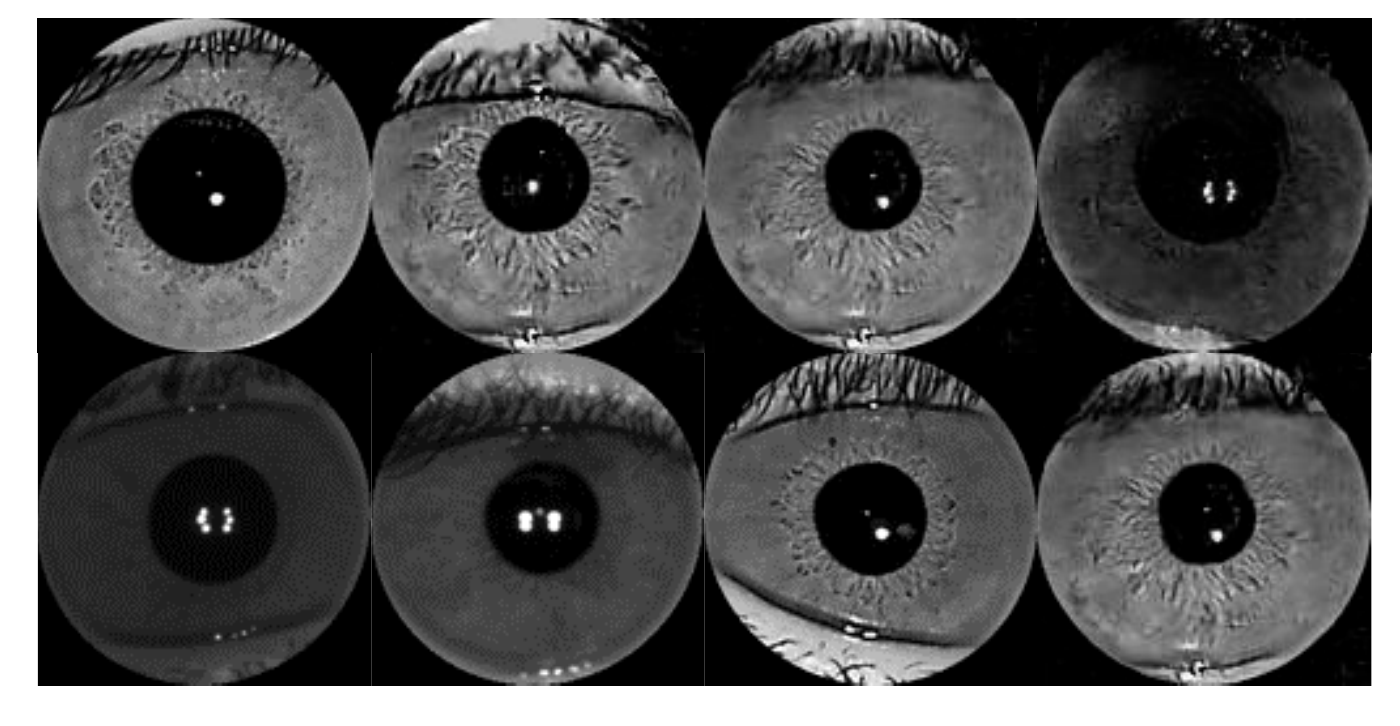}
 \caption{A mixture of real and synthetic iris images generated from the proposed iDCGAN framework are shown above. We encourage the readers to identify which of these iris images are real and synthetic. The solution is shown in Figure \ref{fig:solution} on page 6.}
 \label{fig:motivation}
\end{figure}

\textit{Beauty lies in the iris of the beholder! Some of the iris images in Figure \ref{fig:motivation} are not real  iris images. Can you identify which ones have been generated synthetically? } \\

Ratha et. al. \cite{ratha2001enhancing} presented several avenues of attack on a biometric system and suggested different steps to mitigate such attacks. One of these avenues is through presentation attacks at sensor level which can be used both for identity impersonation and identity evasion. The other potential point of attack in a biometric system is the transmission channel between the sensing device and the feature extraction module \cite{ratha2001enhancing}. A man-in-the-middle attack on this channel can be utilized to replace the original image with a new synthetic image before the template extraction process. The consequences of such an attack maybe wide-ranging as an individual may enrol with different identities and avail facilities associated with the unique ID multiple times.
 
\begin{figure}[t]
 \centering
 \includegraphics[width=0.9\linewidth]{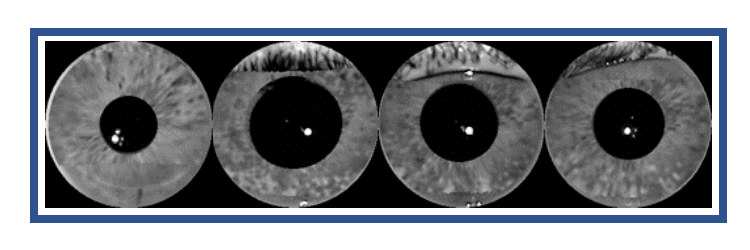}
 \caption{Sample images from Synthetic DataBase \cite{Galbally2013}.} 
 \label{fig:ExistingSamples}
 \vspace{-5mm}
\end{figure}

Presentation attacks on iris modality such as textured contact lenses \cite{kohli2013revisiting,yadav2014unraveling}, synthetically generated iris \cite{Galbally2013}, and print attacks \cite{gupta_print2014} have been explored in the literature. The idea of generating synthetic iris images was initially introduced by Cui et al. \cite{cui2004} with the intention of increasing the number of available iris images for developing iris recognition algorithms. They employed principal component analysis and super-resolution techniques to create new images for iris synthesis. Shah and Ross \cite{shah2006generating} employed Markov Random Field to generate initial texture of the iris images followed by embedding iris features such as radial and concentric furrows to create the final synthetic iris image.  Zuo et al. \cite{zuo2007} developed an anatomy based model to create new irises similar to real-world iris images. Galbally et al. \cite{Galbally2013} reconstructed synthetic iris images from the feature template to successfully match the original genuine iris image.  Figure \ref{fig:ExistingSamples} shows sample synthetic iris images from Synthetic DataBase (SDB) by Galbally et al. \cite{Galbally2013}. It is seen that these images do not resemble real iris images and appear \textit{fake}.

\begin{figure*}[t!]
    \centering
        \centering
        \includegraphics[width=1.\linewidth]{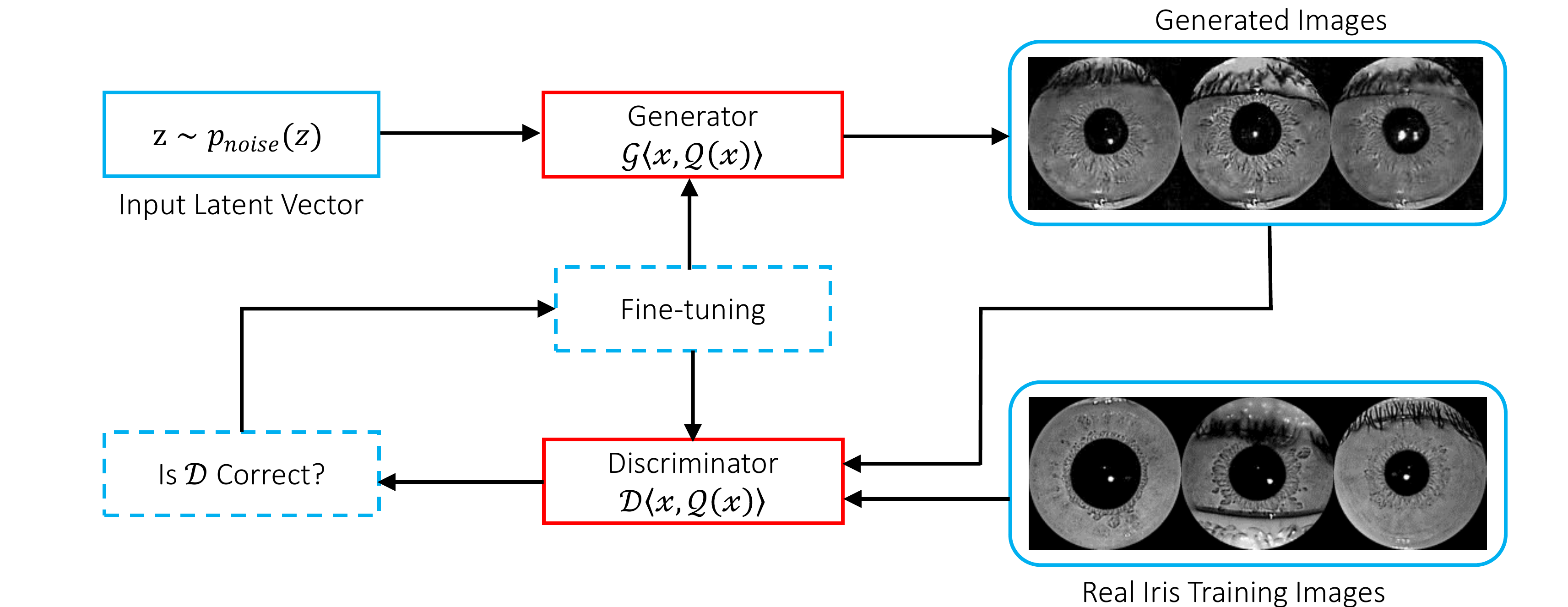}
    \caption{Illustrating the proposed iDCGAN framework for generating synthetic iris images.}
    \label{fig:framework}
\end{figure*}

In this paper, we propose a new iris presentation attack by synthesizing iris images through deep convolutional generative adversarial network. Recently, improvements in techniques such as generative adversarial networks \cite{goodfellow2014generative} and variational autoencoders \cite{kingma2013auto} have provided a breakthrough in generative modeling. These approaches have paved the path for generating realistic looking synthetic images for different applications. In this research, we have proposed a novel synthetic iris image generation method using generative adversarial network and demonstrated that it can attack iris recognition systems. The major contributions of this paper are:

\begin{enumerate}
\itemsep-0.7mm
\item A novel domain specific generative adversarial network (GAN) named as iDCGAN for generating synthetic iris images is proposed. We adapt deep convolutional generative adversarial network by utilizing iris quality assessment for synthesizing \textit{realistic looking} iris images. 

\item Analysis is performed using quality score distributions of real and synthetically generated iris images to understand the effectiveness of the proposed approach. We also demonstrate that synthetically generated iris images can be used to attack existing iris recognition systems. 

\item Evaluation using state-of-the-art iris presentation attack detection algorithm is performed to ascertain its efficacy in distinguishing these synthetically generated images from real images. 
\end{enumerate}

\section{Generative Adversarial Network for Iris Image Generation}
In this research, we adapt generative adversarial network for synthesizing realistic iris images to propose iris Deep Convolutional Generative Adversarial Network (iDCGAN). Figure \ref{fig:framework} shows the steps involved in the proposed approach.

\subsection{Generative Adversarial Network}
Goodfellow et al. \cite{goodfellow2014generative} introduced the concept of generative adversarial networks (GANs) where the generative model is pitted against an adversarial \textit{discriminator} to generate representations which cannot be differentiated by the discriminator. The aim of the \textit{generator} is to learn the probability distribution of the input data perfectly enough to \textit{fool} the discriminator. 

Let $\mathbf{x}$ be the input data which has a true probability distribution $p(\mathbf{x})$. Let $\mathcal{G}$ be the generative network which takes an input latent vector $\mathbf{z}$, drawn from a noisy probability distribution $p_{noise}(\mathbf{z})$ and outputs a new image $\mathbf{\bar{x}}$. Then, the discriminator network $\mathcal{D}$ has to discern if the input image, randomly chosen from $\mathbf{x}$ or $\mathbf{\bar{x}}$, is generated from the true probability distribution $p(\mathbf{x})$ or not. The two models are trained using a minimax objective and the loss function $L$ is shown in Eq. \ref{eq:GAN}.

\begin{equation}
   \begin{split}
	L &= \min_{\mathcal{G}} \max_{\mathcal{D}} \mathbb{E}_{ \mathbf{x} \sim p( \mathbf{x})} [ log (\mathcal{D}(\mathbf{x})) ] \\ 
      &\qquad +  \mathbb{E}_{\mathbf{z} \sim p_{noise}(\mathbf{z})} [ 1 - log (\mathcal{G}(\mathcal{D}( \mathbf{z} )) ] 
   \end{split}
   \label{eq:GAN}
\end{equation}

A number of variants of GANs have been introduced such as conditional GANs \cite{MirzaO14}, Laplacian GANs \cite{denton2015deep}, and InfoGANs \cite{chen2016infogan}. These variants have been successfully utilized in image inpainting \cite{YehCLHD16}, style transfer \cite{WangG16}, and super-resolution \cite{LedigTHCATTWS16} applications. Recently, Shrivastava et al. proposed SimGAN  \cite{ShrivastavaPTSW16} which uses a refiner network to improve appearance of synthetically generated eye images to make them indistinguishable from real eye images.

\begin{figure}[t]
 \centering
 \includegraphics[width=0.9\linewidth]{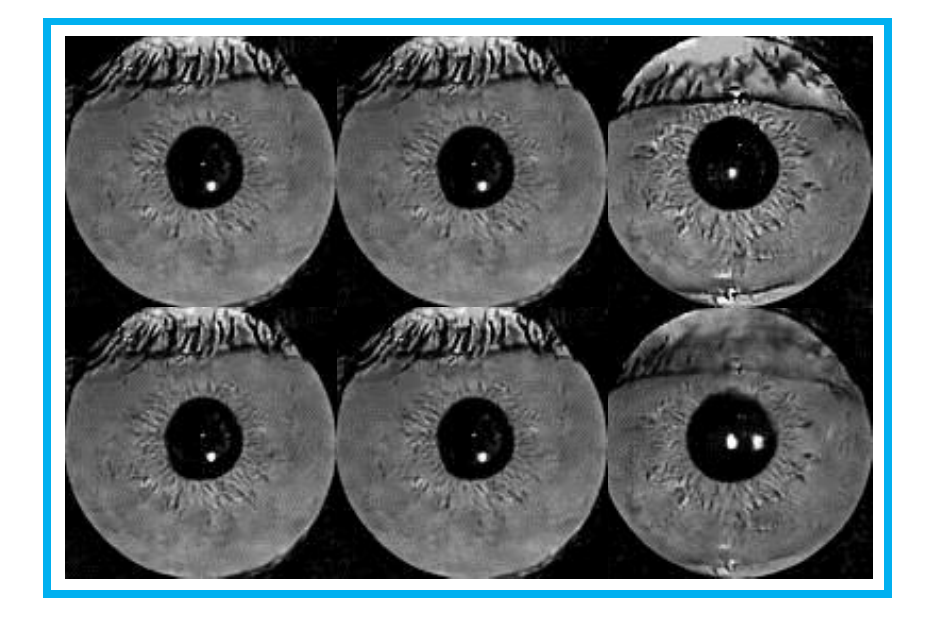}
  \vspace{-4mm}
 \caption{Sample synthetic iris images generated from the proposed iDCGAN framework. } 
 \label{fig:collage}
 \vspace{-3mm}
\end{figure}

\subsection{Proposed iDCGAN for Iris Image Synthesis}

Radford et al. \cite{RadfordMC15} introduced deep convolutional generative adversarial networks (DCGAN) for unsupervised learning of features by utilizing convolutional neural networks as the generator and discriminator network. They also applied constraints on architectural topology of convolutional neural networks in the generator and discriminator networks for stable training. Specifically, pooling functions were replaced with strided convolutions which allowed the resultant network to learn its own spatial upsampling. Additionally, the fully connected layers at the top of convolutional neural networks were removed and batch normalization was utilized for improving model stability by normalizing each unit to have zero mean and unit variance.

In this paper, we propose an extension to DCGAN by utilizing domain (iris) specific knowledge. The new generative adversarial network is termed as iDCGAN (iris Deep Convolutional Generative Adversarial Network). Similar to the idea of conditional GANs \cite{MirzaO14}, it uses auxiliary information of iris quality to improve the performance of both discriminator and generator deep convolutional networks. 

In any iris recognition system, iris image quality assessment is an integral step as the quality of iris images can greatly impact the performance of iris recognition. It has been ascertained that different artifacts such as occlusion, off-gaze direction, motion blurriness, and specular reflection can affect iris recognition performance \cite{kalka2006image,vatsa2008}. Thus, incorporating quality metrics in generative adversarial network can improve the synthesis process. Eq. \ref{eq:iDCGAN} shows the objective function of the proposed iDCGAN framework. 

\begin{equation}
   \begin{split}
	L &= \min_{\mathcal{G}} \max_{\mathcal{D}} \mathbb{E}_{ \mathbf{x} \sim p( \mathbf{x})} [ \log (\mathcal{D} ( \langle \mathbf{x} , Q( \mathbf{x}) \rangle)) ] \\ 
      &\qquad +  \mathbb{E}_{ \mathbf{z} \sim p_{noise}( \mathbf{z})} [ 1 - \log (\mathcal{G}(\mathcal{D}( \langle \mathbf{z} , Q(\mathbf{z} \rangle ))) ] 
   \end{split}
   \label{eq:iDCGAN}
\end{equation}

\noindent where, $Q(\mathbf{x})$ is a quality evaluating function that takes an input iris image and assigns a corresponding quality score. Thus, in the proposed iDCGAN framework the generator network $\mathcal{G}$, spawns new images of iris conditioned on high quality scores. 

The input latent vector is generated from a noisy distribution $p(\mathbf{\mathbf{z}})$. This is provided as input to the generator network, where the generator generates iris images according to the learned representations. Quality assessment of the iris images created by the generator $\mathcal{G}$ is performed. The quality of the iris images in the first quartile are removed from the set to be passed to the discriminator network $\mathcal{D}$. Similar to the above step, the real iris image input to the discriminator network $\mathcal{D}$ is filtered such that the training set contains iris images whose quality scores are above the first quartile. The new samples are continously generated to train the proposed iDCGAN generator and discriminator. Figure \ref{fig:collage} showcases sample iris images generated from the proposed iDCGAN framework.

\subsection{Implementation Details}
Three existing real iris image databases are utilized and combined together to form the training set for the proposed iDCGAN framework:

\begin{description}
\item [IIITD Contact Lens Database] \cite{yadav2014unraveling} This database consists of iris images of 101 subjects. The database includes iris images of subjects with and without contact lens. For training the proposed iDCGAN, only the real images (without contact lens) belonging to these subjects are chosen. \vspace{-1mm}
\item [IIT Delhi Iris Database] \cite{Kumar20101016} This database consists of real iris images pertaining to 224 subjects. \vspace{-1mm}
\item [MultiSensor Iris Database] \cite{kohli2016} Iris images of 547 subjects collected in multiple sessions are utilized for training the proposed iDCGAN framework.
\end{description}

\begin{figure*}[!t]
    \centering
    \begin{subfigure}[t]{0.3\textwidth}
        \centering
        \includegraphics[width=1.05\textwidth]{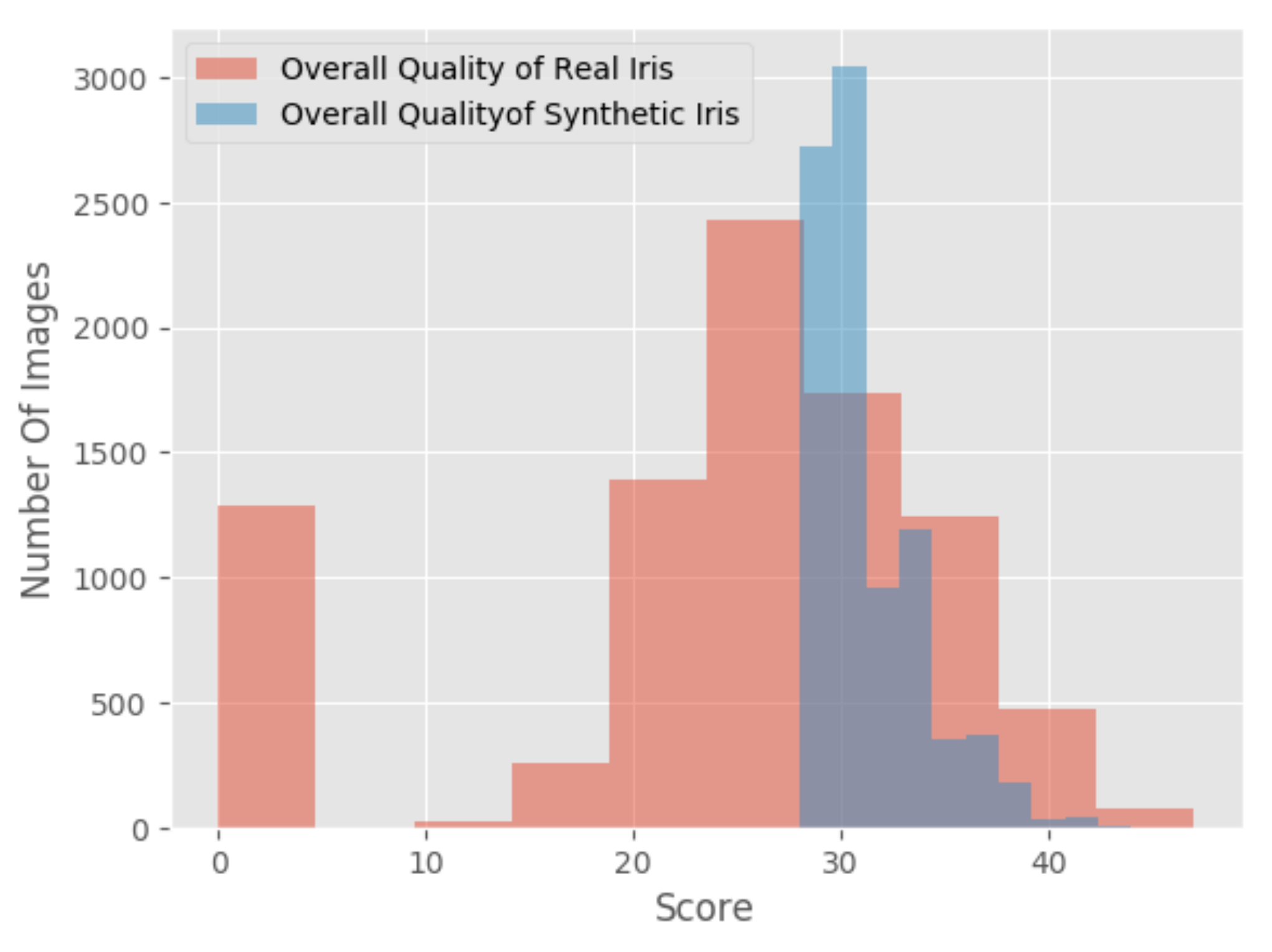} 				
        \caption{Overall Quality}
        
    \end{subfigure}%
    ~ 
    \begin{subfigure}[t]{0.3\textwidth}
        \centering
        \includegraphics[width=1.05\textwidth]{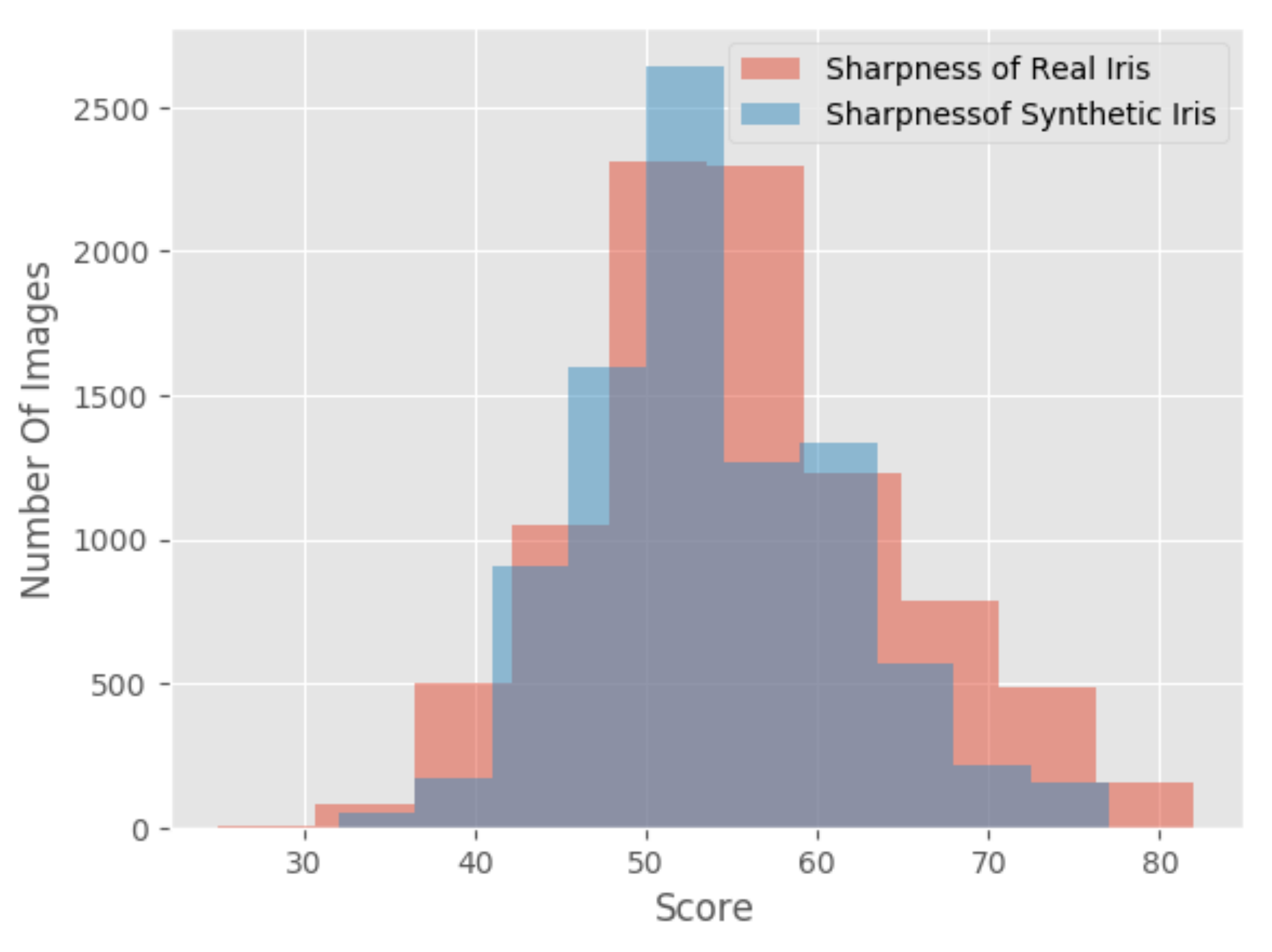}
        \caption{Image Sharpness}
    \end{subfigure}%
    ~
    \begin{subfigure}[t]{0.3\textwidth}
        \centering
        \includegraphics[width=1.05\textwidth]{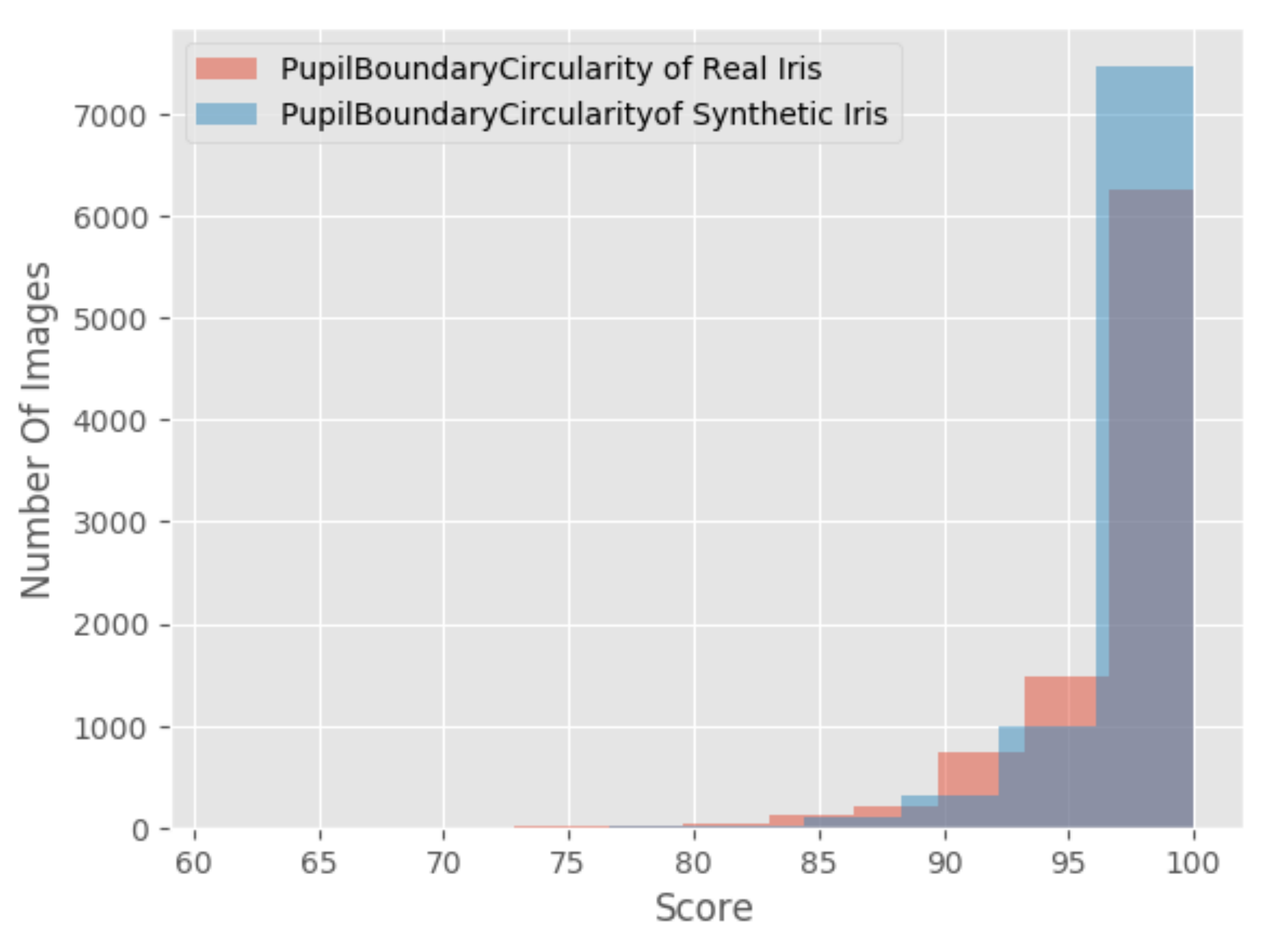}
        \caption{Circularity of Pupil Boundary}
    \end{subfigure}%
    
     \begin{subfigure}[t]{0.3\textwidth}
        \centering
        \includegraphics[width=1.05\textwidth]{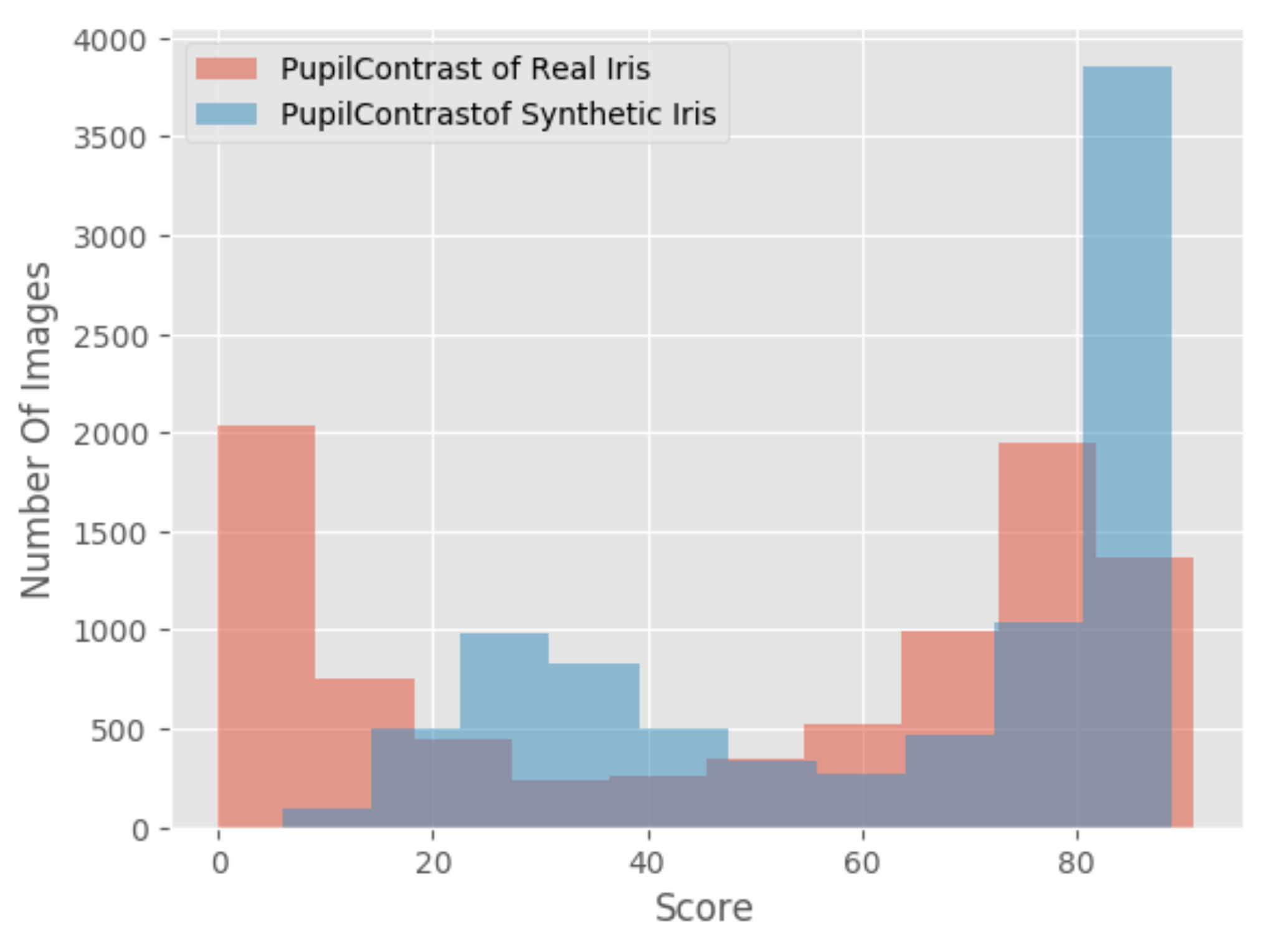}
        \caption{Pupil Contrast}
    \end{subfigure}%
    ~ 
    \begin{subfigure}[t]{0.3\textwidth}
        \centering
        \includegraphics[width=1.05\textwidth]{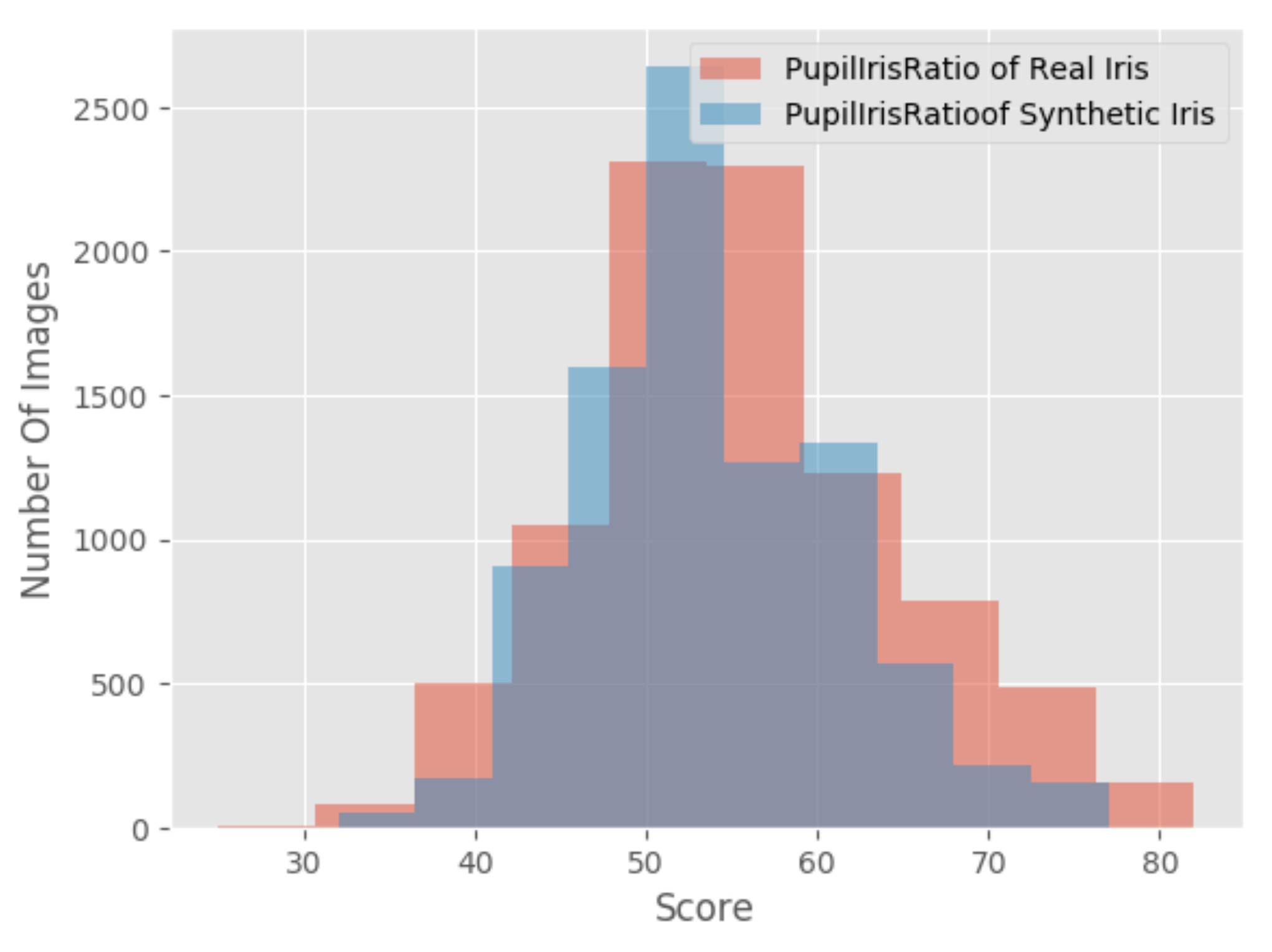}
        \caption{Pupil-to-Iris Ratio}
    \end{subfigure}%
    ~
    \begin{subfigure}[t]{0.3\textwidth}
        \centering
        \includegraphics[width=1.05\textwidth]{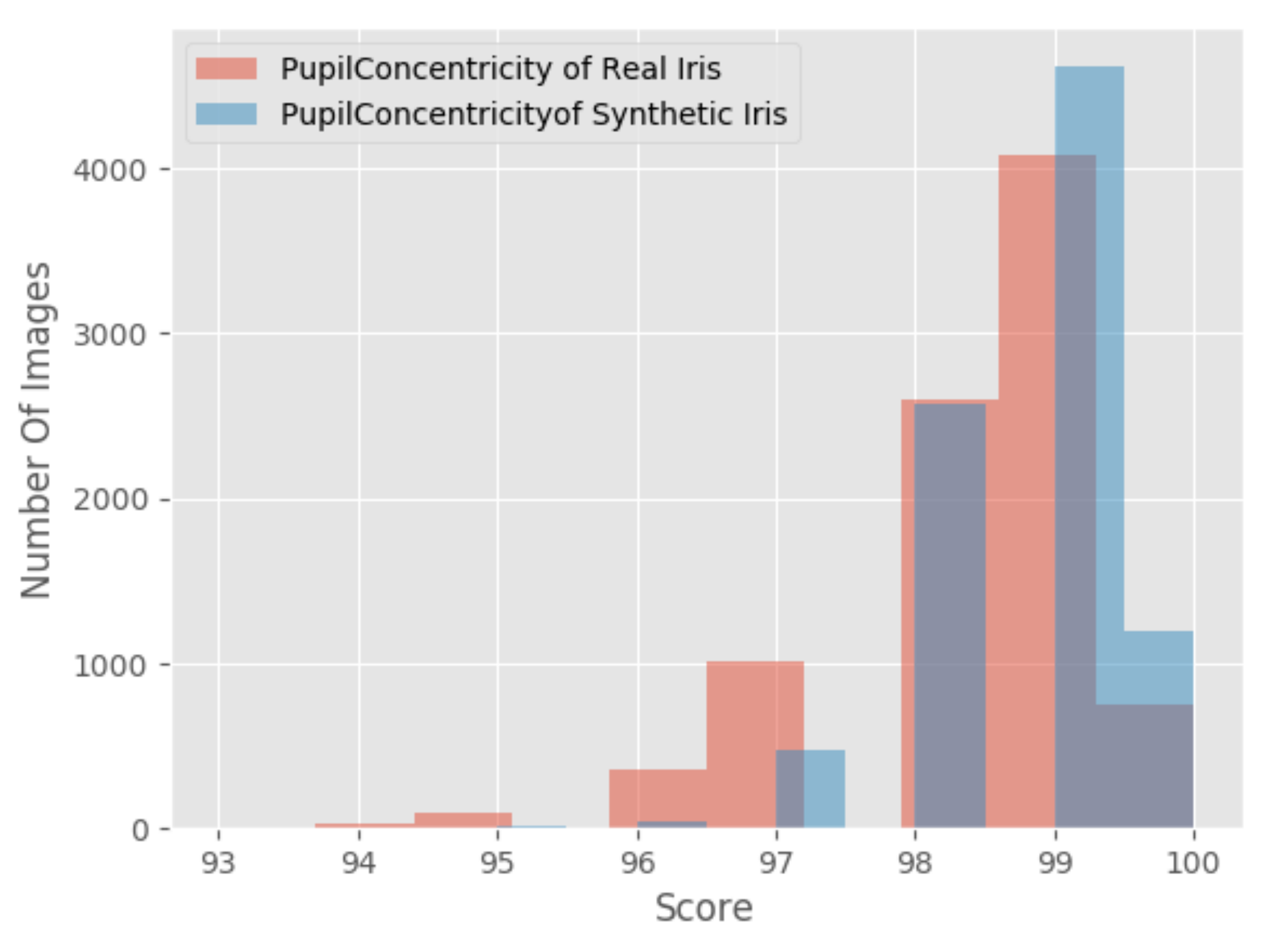}
        \caption{Pupil Concentricity}
    \end{subfigure}%
    \caption{Distribution of various quality metrics highlights the overlap between real and synthetically generated iris images.}
    \label{fig:quality_param}
\end{figure*}

The input iris images are segmented so that only the iris and pupil regions are considered as input to the iDCGAN framework. The framework is implemented in Python language utilizing the TensorFlow library\footnote{https://www.tensorflow.org}. Both the generator and discriminator networks are deep convolutional neural networks. The discriminator network consists of four convolutional layers with kernel size of $5 \times 5$ and strides of 2, batch normalization and leaky rectified units.The generator network consists of four strided transposed convolutional layers with kernel size of $5 \times 5$ and strides of 2, batch normalization and rectified units. The size of the final synthetic iris images is $128 \times 128$. A learning rate of 0.0002 and Adam optimizer are utilized to train the proposed iDCGAN. 

\section{Analysis of Synthetically Generated Iris Images}
The synthetic iris images produced by the proposed iDCGAN framework are evaluated with respect to their similarity with real iris images as well as their ability to attack the iris recognition systems. For this, two experiments are conducted which are described below.

\subsection{Analysis using Iris Quality Metrics}
The iris images generated using the proposed iDCGAN framework are compared with real iris images and are evaluated with respect to different quality score metrics. The quality metrics can evaluate factors such as sharpness of generated images, shape and concentricity of pupil and iris. 

\subsubsection{Experimental Protocol}
The objective of this experiment is to determine the quality of the synthetically generated iris images and compare the quality score distribution with real iris images. Using the combined training set described above, 8,905 real iris images are selected. This is followed by generating an equal number of synthetic iris images using the proposed iDCGAN framework. Bharadwaj et al. \cite{bharadwaj2014biometric} described that the quality of iris images can be categorized into image based and biometric modality based quality measures. Using VeriEye, several image specific and biometric modality specific quality scores are computed. These quality score metrics are described in ISO/IEC 29794-6 standards \cite{iso}. The following quality score metrics are employed for the analysis purposes:
\begin{itemize}
\item Pupil boundary circularity: This parameter represents the circularity of the iris-pupil boundary. It is calculated as $\left(2 * \sqrt{ \pi \times \text{pupil area} } \ \right) / \left( \text{pupil perimeter} \right) $ .
\item Pupil contrast: The contrast value at the boundary of iris and pupil is an important parameter for successful iris segmentation. It is computed as the mean of differences in grayscale values at left and right end of iris-pupil boundary.
\item Pupil-iris ratio: This quality measure signifies the amount of dilation or constriction in the pupil.  
\item Pupil concentricity: This parameter measures the corresponding concentricity between the iris and the pupil. It is calculated as $\sqrt{(X_{pupil} - X_{iris})^{2} + (Y_{pupil} - Y_{iris})^{2}}/ IrisRadius$ where $X$ and $Y$ represent the coordinates of the iris and pupil. 
\item Sharpness: The sharpness of the image parameter is examined to understand the magnitude of defocus in the input iris image. This is calculated using Daugman's focus score \cite{daugman2004iris}.
\item Overall quality: The overall quality score of the iris image represents the comprehensive biometric quality of the presented iris sample. We have utilized output quality score generated from VeriEye.
\end{itemize}

\subsubsection{Results and Analysis}
Figure \ref{fig:quality_param} showcases the distributions of the above mentioned quality parameters pertaining to real iris images and synthetically generated iris images. We observe that the quality measurements of the synthetically generated images follow similar trends to the real iris images. The analysis of the quality metrics can be categorized as follows:

\textbf{Image based Quality:} The sharpness score is an image based quality metric. It is observed that there is a significant overlap between the histograms of sharpness observed in real iris images and synthetically generated iris images. The $\chi^{2}$ distance between the sharpness quality histograms is 1.07 which is relatively low\footnote{Lower $\chi^{2}$ distance values signify very close match.}. Similarly, pupil contrast parameter represents contrast difference in a specific region of interest in the image. The $\chi^{2}$ distance between the pupil contrast histogram is 4.02. It can be observed that the pupil contrast of synthetically generated images is skewed on the higher side as compared to the pupil contrast of real iris images.  Thus, larger number of synthetically generated iris images using the proposed iDCGAN framework have higher pupil contrast score as compared to real iris images. 

\textbf{Biometric based Quality:} The pupil-iris ratio, pupil boundary circularity, and pupil concentricity are measures of the iris biometric modality. We observe that there is a significant overlap between the distribution of pupil-iris ratio, pupil concentricity and pupil boundary which is also confirmed by the $\chi^{2}$ distance of 1.07, 0.04 and 0.34, respectively.

\textbf{Overall Quality:} The quality of the synthetically generated iris images is skewed on the higher side and is different from the quality of the real iris images in the combined training set.  The generator network in the proposed iDCGAN framework is trained to discard iris images that are not of good quality. Therefore, it has generated high quality synthetic images.

The comparative analysis of these quality score metrics indicates that the synthetically generated iris images very closely resemble the real iris images. 

\subsection{Synthetic Iris as Presentation Attack}
The objective of the proposed iDCGAN framework is to generate iris images which appear \textit{real}. Due to the realistic appearance of these synthetic iris images, they can be used as an attack on any iris recognition system. In this experiment, we utilize VeriEye \cite{verieye} to examine if a commercial iris recognition matches these synthetic images to real iris images. The results of this experiment are utilized to establish that the output images from the proposed iDCGAN framework can act as an iris presentation attack.  

\subsubsection{Experimental Setup}
The goal of this experiment is to compute iris recognition scores between gallery and probe sets to evaluate the impact of synthetically generated iris as presentation attacks. 
For this iris recognition experiment, real genuine, real impostor, and synthetic impostor pairs are created using 8,905 real iris images and 8,905 synthetic iris images. The match scores obtained by matching these pairs are analyzed and the results are presented below.

\subsubsection{Results and Analysis}
These real genuine and synthetic impostor scores are analyzed to observe the impact of synthetically generated iris images on the performance of VeriEye. Upon minimizing the synthetic iris false accept to 0\%, we observe that 15.2\% of real iris genuine scores are misclassified as impostors. On the other hand, minimizing the real iris false reject to 0\% leads to synthetic false accept rate of 67.66\%. This showcases that the synthetically generated images adversely affect iris recognition and can pass through the recognition system based on the chosen permissible error threshold.

Interestingly, we observe that all the synthetically generated iris images are encoded by VeriEye and templates are created for every image. A denial of service attack can easily be executed on an iris recognition system by sending such synthetically generated iris images as input. These results validate that the realistic-looking synthetically-generated iris outputs from the proposed iDCGAN framework can be potentially used for iris presentation attack.

\begin{figure}[t]
 \centering
 \includegraphics[width=1.01\linewidth]{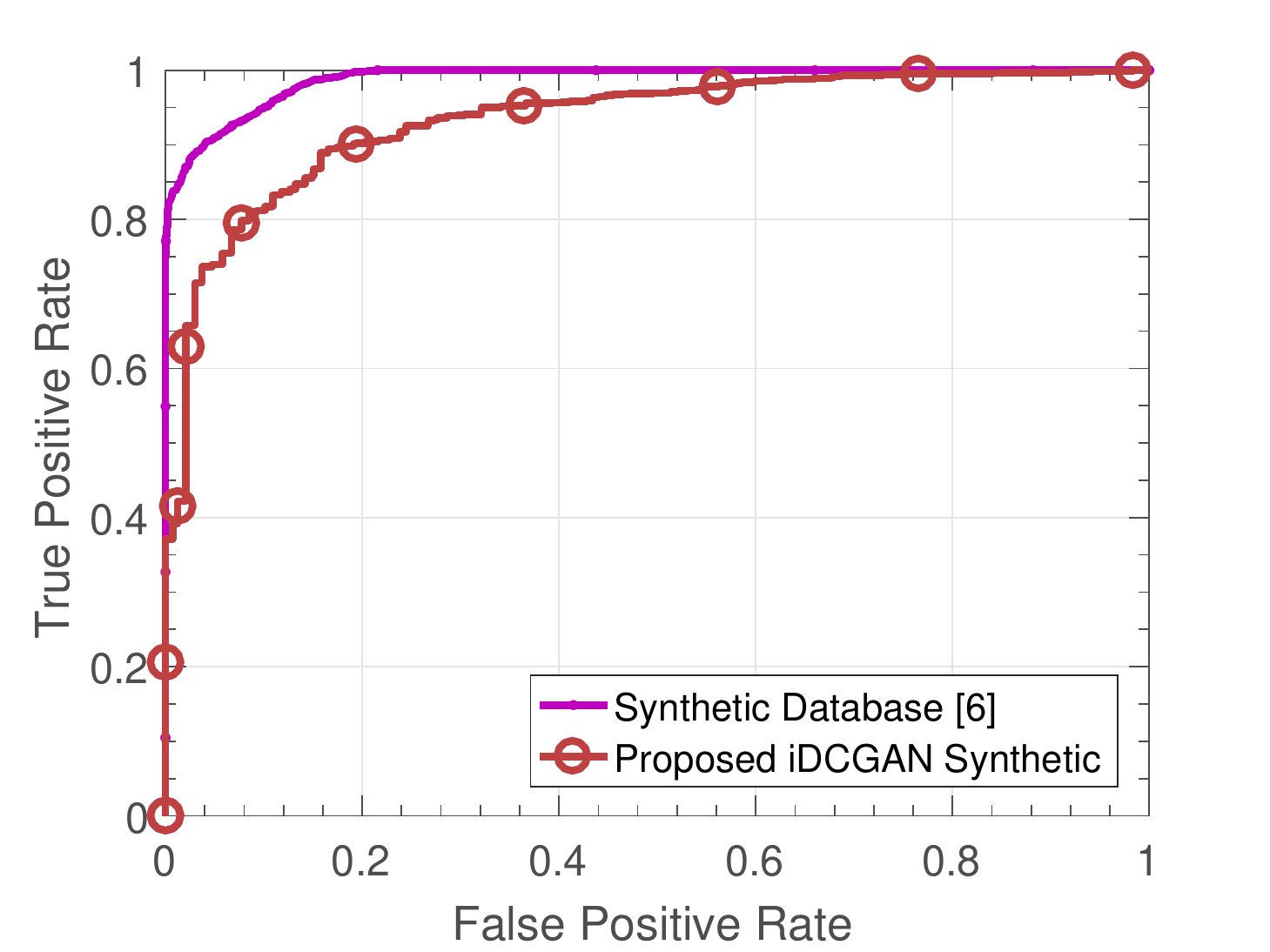}
 \caption{Performance of presentation attack detection using DESIST on images from the Synthetic DataBase \cite{Galbally2013} and the proposed iDCGAN synthetic images. } 
 \label{fig:desistROC}
 \vspace{-5mm}
\end{figure}
\section{Iris Presentation Attack Detection on iDCGAN Generated Iris Images}
The key results of the previous section illustrate that the synthetically generated iris images from the proposed iDCGAN framework can be effectively deployed in iris presentation attacks. Hence, it is important to develop accurate iris presentation attack detection (PAD) algorithms which can distinguish such synthetic iris images from real iris images. In this section, we present baseline results of state-of-the-art PAD algorithm, DESIST \cite{kohli2016}. 

\subsection{Experimental Protocol}
In this experiment, we analyze the performance of DESIST PAD algorithm for detecting synthetically generated iris images. 
To showcase that the synthetically generated iris images using the proposed iDCGAN framework are strong adversary as compared to existing synthetic iris images, we utilize SDB \cite{Galbally2013}. 
SDB comprises 2,100 synthetic iris images. Equal number of real iris images and iris images that are synthetically generated from the iDCGAN approach, are utilized for experimental evaluation. In this experiment, five-fold cross validation is performed with unseen training and testing samples. Multi-order Zernike moments and local binary pattern with variance (LBPV) features are extracted to provide input to the DESIST framework for classifying iris images as real or synthetic using neural network as the classifier.

\subsection{Results}

The results of the presentation attack detection using DESIST are presented in Figure \ref{fig:desistROC}. 
 Iris PAD accuracy on the synthetically generated iris images using the proposed iDCGAN framework is 85.95\% with equal error rate (EER) of 14.19\%. PAD performance of DESIST on SDB  is 92.17\% with an EER of 7.09\%. We observe that EER by DESIST on SDB is approximately 2 times higher than the EER obtained with iDCGAN generated images. As discussed in the previous sections, the iris image quality scores of the realistic appearing synthetically generated samples are closer to the real-world samples and hence, it is difficult for the DESIST model to discriminate between the samples of the real iris and presentation attack iris classes.

\begin{figure}[t]
 \centering
 \includegraphics[width=1.0\linewidth]{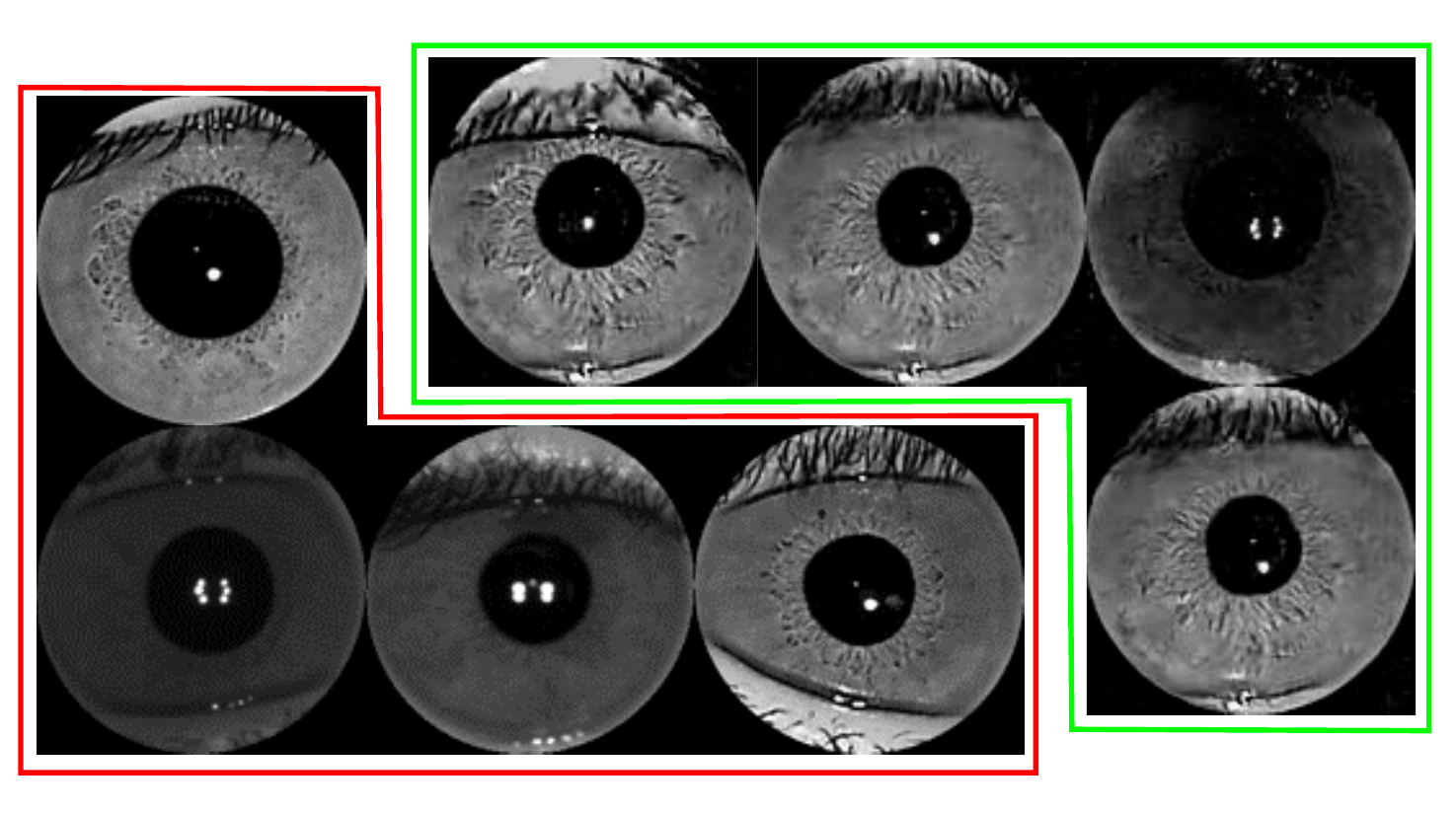}
 \caption{Marked real iris and synthetically generated iris images using the proposed iDCGAN framework. Iris images inside the red border are real iris images and the remaining iris images inside the green border are synthetically generated images.} 
 \label{fig:solution}
 \vspace{-5mm}
\end{figure}

\section{Conclusion}
The answer to the question posed in Figure \ref{fig:motivation} is shown in Figure \ref{fig:solution}. The iDCGAN framework incorporates iris domain specific knowledge in the form of quality metric to generate high quality iris images. It is observed that the distributions of quality parameters described for a biometric sample for the synthetically generated iris images are similar to that of real iris images, thus, establishing the similarity between real and synthetically generated images. We also demonstrate the probability of a successful presentation attack by utilizing these synthetically generated iris images. Finally, state-of-the-art presentation attack detection framework, DESIST is applied to distinguish synthetically generated iris images from real images. It is observed that the synthetically generated iris images from the iDCGAN framework are more challenging to be detected by DESIST compared to existing synthetic iris database. This paper also highlights the need to develop accurate iris presentation attack detection algorithms that can adapt to newer types of attacks.   

\section{Acknowledgement}
The authors gratefully acknowledge the support of NVIDIA Corporation for the donation of the Tesla K40 GPU for this research.

{\small
\bibliographystyle{ieee}
\bibliography{synthetic_irisGAN}
}

\end{document}